\documentclass{article}

\usepackage[sglblindworkshop, final]{neurips_2026}

\workshoptitle{AI at Scale for Clinical Impact (ASCI): Cancer Pathology Foundation Models}
\usepackage{pgfplots}
\pgfplotsset{compat=1.18}
\usepackage[utf8]{inputenc}
\usepackage[T1]{fontenc}
\usepackage{hyperref}
\usepackage{url}
\usepackage{booktabs}
\usepackage{amsmath}
\usepackage{amsfonts}
\usepackage{amssymb}
\usepackage{nicefrac}
\usepackage{microtype}
\usepackage{xcolor}
\usepackage{graphicx}
\usepackage{multirow}
\usepackage{tikz}
\usetikzlibrary{arrows.meta,positioning,fit,backgrounds,calc}

\title{Input- and Feature-Space Correction\\
for Cross-Hospital IHC Grading}

\author{%
  Jongwoon Lee$^{1}$ \quad Jhair Gallardo$^{2}$ \quad
  Christopher Kanan$^{1}$ \\[6pt]
  \small
  $^{1}$Department of Computer Science, University of Rochester, Rochester, NY\\
  $^{2}$Rochester Institute of Technology, Rochester, NY\\[4pt]
  }

\begin{document}

\maketitle

\begin{abstract}
Pathology foundation models (FMs) are trained predominantly on hematoxylin and
eosin (H\&E) images but are increasingly used as frozen feature extractors for
immunohistochemistry (IHC), where cross-hospital differences can impair
grading. However, systematic comparisons of correction methods for cross-hospital IHC grading remain limited. To address this gap, we benchmark label-free transformations applied to input images
before frozen-FM encoding and to features after frozen-FM encoding using UNI and
Virchow2, three IHC markers, and six directed cross-hospital transfers,
yielding 12 settings.
Among the evaluated input-space methods, our proposed method FIX achieved the largest macro-F1 improvement over no correction, outperforming well-known stain normalization techniques. Among feature-space methods, a simple feature-wise standardization achieved the best improvement over no correction, outperforming CORAL.
Under patient-averaged tile-level evaluation, combining FIX with
feature-wise standardization improved macro-F1 and quadratic weighted kappa
(QWK) by 0.181 and 0.208, respectively, and reduced ordinal error by 0.209
relative to no correction. These results show that
input- and feature-space transformations can provide complementary benefits
for cross-hospital IHC grading with frozen pathology FMs.
\end{abstract}

\section{Introduction}
\label{secc:intro}

IHC visualizes the tissue expression of specific
proteins and supports biomarker assessment in precision oncology. HER2 IHC
grading, for example, contributes directly to treatment selection
~\citep{wolff2023her2,modi2022destiny}. Although deep learning may improve the
accuracy and consistency of IHC assessment~\citep{wu2023her2ai}, a model
trained at one institution can lose performance at another because of
differences in tissue processing, staining protocols, and scanners. These
site-specific differences persist in both digital pathology images and their
learned representations~\citep{tellez2019stain,vahadane2016structure,
howard2021site,huang2025flex,thiringer2026scanner}.

This problem is particularly relevant to pathology FMs because
these models are trained predominantly on H\&E images
while increasingly used as frozen feature extractors for IHC. Consequently,
a classifier trained for IHC grading on frozen-FM embeddings from one institution may
encounter substantial input and feature shifts when transferred to another
institution.
To address this problem, label-free corrections can act at different stages of the pipeline.
Input-space methods modify images before FM encoding, where established examples
include Reinhard color transfer and Macenko stain normalization
~\citep{reinhard2001,macenko2009}. Feature-space methods instead modify the
features after FM encoding, including methods such as moment alignment and covariance alignment~\citep{sun2016coral,li2018adabn}.  However, these approaches have been developed and evaluated
primarily for H\&E images, and their relative performance remains unclear for
cross-institutional IHC grading.

We address this gap through a controlled comparison of label-free input- and
feature-space corrections for cross-institutional IHC grading with frozen
pathology FMs. The evaluated input-space methods include Reinhard
normalization, Macenko normalization, PCA projection, and our proposed component-wise
input correction method that we call FIX. The feature-space methods include
mean shift (MS), feature-wise standardization (FS), and CORAL. We evaluate these methods using
UNI~\citep{chen2024uni} and Virchow2~\citep{zimmermann2024virchow2} across 3
IHC markers and 6 directed cross-institutional transfers, yielding 12
settings. We also evaluate combinations of input- and feature-space methods
to determine whether corrections on opposite sides of the frozen encoder
provide complementary benefits.

Our contributions are threefold:
\begin{itemize}
    \item We provide a controlled comparison of label-free input- and
    feature-space transformations for cross-institutional IHC grading with
    frozen pathology FMs.

    \item We introduce FIX, an input-space transformation based on source-patch PCA basis and component-wise input correction. Among the evaluated standalone input-space
    methods, FIX achieved the largest mean macro-F1 improvement.

    \item We show that input- and feature-space transformations can provide
    complementary benefits, where FIX + FS achieved the highest mean performance
    among the evaluated combinations.
\end{itemize}

\section{Related Work}

\paragraph{Pathology FMs.}
Pathology FMs are self-supervised encoders trained on large slide collections
and commonly used as frozen feature extractors. We evaluate UNI, trained
predominantly on H\&E tiles, and Virchow2, whose larger pretraining corpus also
includes IHC slides~\citep{chen2024uni,zimmermann2024virchow2}.

\paragraph{Label-free cross-institutional correction.}
Input-space methods such as Reinhard and Macenko modify color or stain
distributions before encoding~\citep{reinhard2001,macenko2009}. Feature-space
methods instead align embedding statistics: MS matches source and target means,
whereas CORAL additionally matches covariance
~\citep{tzeng2014ddc,li2018adabn,sun2016coral}. We compare
these correction stages while keeping the pathology FM frozen.

\paragraph{Gap addressed.}
Cross-institutional adaptation of frozen pathology FMs remains insufficiently
studied for IHC, where marker-specific reagents, antibody clones, staining
platforms, and scanners introduce substantial variation
~\citep{hirsch2017blueprint}. In this work, we study input- and feature-space methods under cross-institutional adaptation across 3 IHC markers and 2 frozen pathology FMs.

\section{Methods}
\label{sec:methods}

We study label-free transfer from a labeled source institution to a target
institution using a frozen pathology FM. We organize the evaluated
transformations according to where they act relative to the encoder.
Input-space methods modify RGB images before encoding, while Feature-space methods modify the
FM output representation. We also evaluate combinations of input- and feature-space
transformations to test whether corrections applied on opposite sides of the
frozen encoder provide complementary benefits.

\begin{figure}[t]
    \centering
    \includegraphics[width=\linewidth]{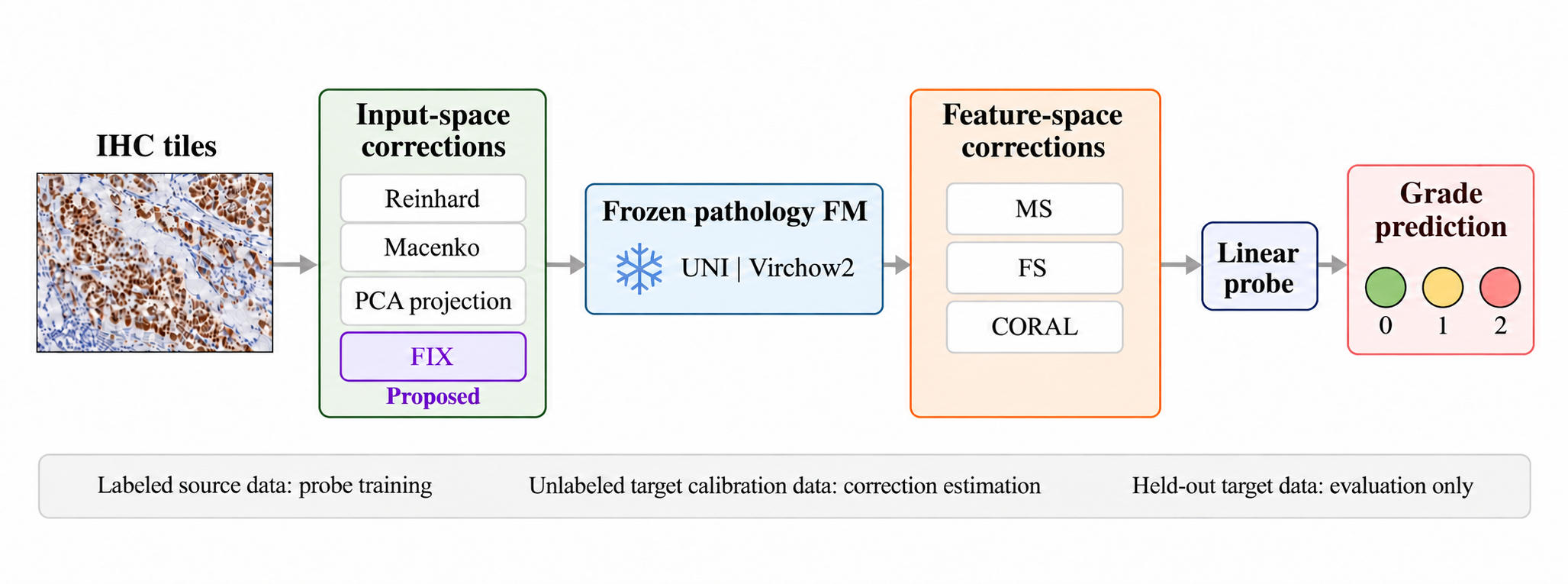}
\caption{\textbf{Overview of the evaluated label-free transformations.}
Input-space methods act before frozen-FM encoding, whereas feature-space
methods act on the resulting embeddings. 
}
    \label{fig:pipeline}
\end{figure}

More formally, let
$\mathcal{D}_s=\{(x_i^s,y_i^s)\}_{i=1}^{n_s}$,
$\mathcal{D}_t^{\mathrm{cal}}=\{x_j^c\}_{j=1}^{n_c}$, and
$\mathcal{D}_t^{\mathrm{eval}}=\{(x_\ell^e,y_\ell^e)\}_{\ell=1}^{n_e}$
denote the labeled source, unlabeled target calibration, and held-out target
evaluation sets, respectively. Here, $x$ denotes an image tile and
$y\in\{0,1,2\}$ its grade. $n_s$, $n_c$, and $n_e$ are the numbers of source,
target calibration, and target evaluation tiles, respectively. The frozen FM
$f_\theta$ maps each tile to a $d$-dimensional feature vector. Input-space methods act on the tile $x$, whereas feature-space methods act on the $d$-dimensional feature vector. The labeled source set $\mathcal{D}_s$ is used for probe training; the unlabeled target calibration set $\mathcal{D}_t^{\mathrm{cal}}$ is used for correction estimation; and the held-out target
evaluation set $\mathcal{D}_t^{\mathrm{eval}}$ is used for the calculation of the final performance metrics. Figure~\ref{fig:pipeline} summarizes the pipeline.

\subsection{Input-space methods} 

\paragraph{Stain normalization.}
Stain normalization reduces inter-institutional color variation by mapping
source images toward color or stain statistics estimated from the unlabeled
target calibration set. Reinhard normalization matches channel-wise LAB color
statistics, whereas Macenko normalization estimates and aligns stain vectors
and concentrations~\citep{reinhard2001,macenko2009}. For each source--target
direction, the reference statistics were pooled from the target calibration
tiles without using evaluation tiles or target grades. The fitted
transformation was applied only to the labeled source tiles used for probe
training; held-out target tiles were encoded without stain normalization.

\paragraph{PCA projection.}
PCA projection reconstructs both source and target RGB tiles using a subset of principal components (PC) fitted using source tiles. We randomly sample patches from source tiles and collect them in
$P\in\mathbb{R}^{N\times D}$, where $N$ is the total number of sampled
source patches and $D=16\times16\times3=768$ is the flattened patch
dimension. Let $\bar{p}$ denote the source-patch mean and
$V_K\in\mathbb{R}^{D\times K}$ the eigenvectors associated with the $K$
largest eigenvalues of the centered source-patch covariance matrix. For a
source or target patch $p$, its coefficients and reconstruction are
\begin{equation}
c=(p-\bar{p})V_K,
\qquad
\widehat{p}=cV_K^\top+\bar{p}.
\end{equation}
We fixed $K=64$, which retains
98.6--99.9\% of the source-patch variance across sources.

To reconstruct a complete tile instead of a single patch $p$, each tile is resized to $224\times224$ pixels and divided into $16\times16$ RGB patches with stride 8.
Because the stride is smaller than the patch size, adjacent patches overlap. After reconstructing each patch,
the patches are returned to their original image locations, and overlapping
pixel values are averaged to reconstruct the tile.

\paragraph{Component-wise input correction (FIX).}
FIX extends source-basis PCA projection by selectively correcting target-patch
coefficients whose distributions differ from those of the source. Both source
and target tiles are reconstructed using the fixed source PCA basis. Source
tiles use PCA reconstruction without coefficient editing, whereas FIX
modifies target coefficients along the fixed source PCA axes before
reconstruction. FIX does not modify the PCA basis vectors themselves. Its
coefficient statistics are estimated using source patches and unlabeled
target calibration patches, without using held-out target evaluation tiles or
target grades.

Having $\bar{p}\in\mathbb{R}^{D}$ as the mean source patch and
$V_K\in\mathbb{R}^{D\times K}$ as the retained source PCA basis, the coefficients for a flattened
patch $p\in\mathbb{R}^{D}$ are:
\begin{equation}
c=(p-\bar{p})V_K,
\qquad
c=(c_1,\ldots,c_K).
\end{equation}

For component $i$, let $c_{m,i}^{s}$ and
$c_{n,i}^{t,\mathrm{cal}}$ denote the coefficients of source patch $m$ and
target calibration patch $n$, respectively, where $N_s$ and $N_t$ are the
corresponding numbers of patches. We calculate the coefficient means and
sample standard deviations separately for the two sets:
\begin{equation}
\mu_{s,i}
=
\frac{1}{N_s}
\sum_{m=1}^{N_s}
c_{m,i}^{s},
\qquad
\mu_{t,i}
=
\frac{1}{N_t}
\sum_{n=1}^{N_t}
c_{n,i}^{t,\mathrm{cal}}.
\end{equation}

\begin{equation}
\sigma_{s,i}
=
\sqrt{\frac{1}{N_s-1}
\sum_{m=1}^{N_s}(c_{m,i}^{s}-\mu_{s,i})^2},
\qquad
\sigma_{t,i}
=
\sqrt{\frac{1}{N_t-1}
\sum_{n=1}^{N_t}(c_{n,i}^{t,\mathrm{cal}}-\mu_{t,i})^2}
\end{equation}

For each PC $i$, we combine the differences in the source and target means and
standard deviations as:
\begin{equation}
W_{2,i}^{2}
=
(\mu_{s,i}-\mu_{t,i})^2
+
(\sigma_{s,i}-\sigma_{t,i})^2.
\end{equation}
where $W_{2,i}^{2}$ is the squared 2-Wasserstein distance between the corresponding
one-dimensional Gaussian distributions~\citep{dowson1982frechet}. We normalize
it by the average within-domain variance to obtain a dimensionless drift
score:
\begin{equation}
r_i
=
\frac{
W_{2,i}^{2}
}{
\tfrac{1}{2}
\left(\sigma_{s,i}^{2}+\sigma_{t,i}^{2}\right)
+\varepsilon_r
},
\qquad
\varepsilon_r=10^{-8}.
\end{equation}

where larger $r_i$ indicates greater source--target discrepancy on component $i$.

For a target patch coefficient $c_i$, FIX first computes
\begin{equation}
u_i=c_i-\mu_{t,i},
\qquad
v_i
=
u_i
\frac{\sigma_{s,i}}
{\sigma_{t,i}+\varepsilon_\sigma},
\qquad
c_i^{\mathrm{align}}
=
v_i+\mu_{s,i},
\end{equation}
where $\varepsilon_\sigma=10^{-5}$ prevents division by zero. Full alignment
matches the target coefficient mean and standard deviation to the
corresponding source statistics. Let $\mathbf{r}=(r_1,\ldots,r_K)$ denote the component-wise drift scores and
$\boldsymbol{\lambda}=(\lambda_1,\ldots,\lambda_K)$ the corresponding PCA
eigenvalues. FIX uses the prespecified gating function $g_i$ to assign a
component-specific correction strength:
\begin{equation}
\alpha_i
=
g_i(\mathbf{r},\boldsymbol{\lambda}),
\qquad
\hat{c}_i
=
(1-\alpha_i)c_i
+
\alpha_i c_i^{\mathrm{align}},
\qquad
0\leq\alpha_i\leq1.
\end{equation}
Here, $\alpha_i=0$ leaves target coefficients on component $i$ unchanged,
whereas $\alpha_i=1$ fully aligns their mean and standard deviation with the
corresponding source statistics. Intermediate values produce partial
alignment. The fixed rule used to compute $\alpha_i$ from the drift scores and
source eigenvalues is provided in Section~\ref{sup:fix_gating}.

The corrected patch is returned to RGB space using the inverse PCA transform:
\begin{equation}
\hat{p}=\hat{c}V_K^\top+\bar{p},
\end{equation}
Once estimated, the same FIX transformation is applied to held-out target tiles without re-estimation. FIX follows the same reconstruction process as PCA projection for a complete tile.

\subsection{Feature-space methods}

\paragraph{Mean-shift (MS).}
MS aligns the labeled source embeddings with the center of the unlabeled
target calibration distribution before classifier training. Let $f_\theta$
denote the frozen FM and let
\begin{equation}
\bar{z}_s
=
\frac{1}{n_s}\sum_{i=1}^{n_s} f_\theta(x_i^s),
\qquad
\bar{z}_t^{\mathrm{cal}}
=
\frac{1}{n_c}\sum_{j=1}^{n_c} f_\theta(x_j^c)
\end{equation}
be the source and target calibration feature means. Each labeled source feature
is translated as
\begin{equation}
\hat{z}_i^{s}
=
f_\theta(x_i^s)-\bar{z}_s+\bar{z}_t^{\mathrm{cal}},
\end{equation}
and remains paired with its original source grade $y_i^s$. The linear probe is
trained on these shifted source features. MS is a
closed-form first-moment alignment in the frozen feature space, unlike
MMD-based approaches that learn representations using a
distribution-matching loss~\citep{tzeng2014ddc}.

\paragraph{Feature-wise standardization (FS).}
FS extends MS by matching both the mean and standard
deviation of each source-feature dimension to those estimated from the
unlabeled target calibration embeddings:
\begin{equation}
\hat{z}_i^{s}
=
\frac{f_\theta(x_i^s)-\bar{z}_s}{\max(\sigma_{s},\varepsilon_s)}\sigma_{t}^{\mathrm{cal}}+\bar{z}_t^{\mathrm{cal}},
\end{equation}
where $\sigma_{s}$ and $\sigma_{t}^{\mathrm{cal}}$ are sample
standard deviations from the source and calibration sets, respectively. $\varepsilon_s = 10^{-8}$ prevents division by zero. FS is related to adaptive
batch normalization through its use of domain-specific means and variances,
but it operates on frozen output embeddings rather than replacing statistics
within the encoder's batch-normalization layers~\citep{li2018adabn}.

\paragraph{CORAL.}
CORAL aligns the covariance of labeled source embeddings with that
estimated from unlabeled target calibration embeddings
~\citep{sun2016coral}. An additional variant of CORAL called Ledoit--Wolf CORAL was also explored. Ledoit--Wolf CORAL applies the full-covariance
alignment using shrinkage covariance estimates to improve numerical stability
in high-dimensional feature spaces~\citep{ledoit2004}.

\section{Experimental Setup}
\label{sec:exp_setup}

\paragraph{Data and tile annotations.}
The available pool contained 7,499 unique tiles. HER2 comprised 1,500 tiles
from 15 patients at Institution A and 1,500 tiles from BCI with no patient identifiers. Nectin-4 comprised 1,500 tiles from each institution. At Institution A,
1,023 of 1,500 tiles were linked to 82 identified patients; patient
identifiers were unavailable for the remaining 477 tiles. At Institution B,
all 1,500 tiles were linked to 120 identified patients. TROP2 comprised 749 tiles from 67 patients at Institution A and 750 tiles from 32 patients at Institution B. Two pathologists independently graded every tile and resolved disagreements by a consensus meeting. For Nectin-4 and TROP2, grade 0 indicated no appreciable expression, grade 1 indicated weak or limited expression, and grade 2 indicated stronger or more extensive expression.
For HER2, original IHC scores of 0 or 1+, 2+, and 3+ were mapped to study
grades 0, 1, and 2, respectively~\citep{wolff2023her2}. For
each marker, each cohort served in turn as the labeled source and the other as
the target. Combined with two frozen FMs, this yielded 12 settings
($2$ encoders $\times$ $3$ markers $\times$ $2$ source--target assignments). 
Additional details of the datasets, including handling of tiles with missing patient identifiers, are included in Section~\ref{sup:data_details}.

\paragraph{Foundation-model representation.}
We compare each of the methods described in Section~\ref{sec:methods} using two FMs: UNI and Virchow2. For UNI, we use the recommended 1,024-dimensional class-token
representation~\citep{chen2024uni}. For Virchow2, we concatenate its
1,280-dimensional class token and the 1,280-dimensional mean patch-token
representation, yielding the recommended 2,560-dimensional
embedding~\citep{zimmermann2024virchow2}.

\paragraph{Probe training.}
For each correction method, encoder, transfer direction, and
calibration--evaluation split, we trained a separate linear probe using only
the labeled source embeddings and their source grades. The source embeddings
were first transformed when required by the evaluated method; for example, MS
and FS aligned the source embeddings using statistics estimated from the
unlabeled target calibration set. Target calibration grades and held-out
target evaluation data were not used for probe training.
The probe was a multinomial logistic-regression classifier implemented with
the L-BFGS optimizer. We used $C=1.0$, balanced class weights, a maximum of
2{,}000 iterations, a convergence tolerance of $10^{-4}$, float64 inputs, and
a fixed random seed of 42. These settings were fixed before the primary
experiments and were not selected using target evaluation performance. After
training, the probe was applied to the corresponding held-out target
evaluation embeddings to calculate the final performance metrics.

\paragraph{Metrics.}
We report tile-level and patient-averaged tile-level macro-F1, quadratic weighted kappa (QWK), and ordinal
error. Macro-F1 gives equal importance to the 3 grades. QWK assigns larger
penalties to errors between more distant grades~\citep{cohen1968kappa}, and
ordinal error is the mean absolute difference between the predicted and
reference grades.
Within each of the 12 settings, we average performance across five
calibration--evaluation splits.

\section{Results}
\label{sec:results}

\begin{table*}[t]
    \centering
    \small
    \setlength{\tabcolsep}{3pt}
    \renewcommand{\arraystretch}{1.1}
\caption{Tile-level macro-F1 for label-free correction methods using UNI.
Values are means over five target calibration--evaluation splits.
Mean denotes the average across the six transfer directions.
Bold indicates the numerically highest value.}
    \label{tab:correction_uni}
    \begin{tabular}{@{}llccccccc@{}}
        \toprule
        \multirow{2}{*}{Method}
        & \multirow{2}{*}{Correction stage}
        & \multicolumn{2}{c}{HER2}
        & \multicolumn{2}{c}{Nectin-4}
        & \multicolumn{2}{c}{TROP2} \\
        \cmidrule(lr){3-4}
        \cmidrule(lr){5-6}
        \cmidrule(lr){7-8}
        & &
        $\mathrm{A}\to\mathrm{BCI}$ &
        $\mathrm{BCI}\to\mathrm{A}$ &
        $\mathrm{A}\to\mathrm{B}$ &
        $\mathrm{B}\to\mathrm{A}$ &
        $\mathrm{A}\to\mathrm{B}$ &
        $\mathrm{B}\to\mathrm{A}$ &
        Mean \\
        \midrule
        No correction                & None                 & 0.356 & 0.564 & 0.412 & 0.549 & 0.817 & 0.699 & 0.566 \\
        \midrule
        Reinhard                     & Input                & 0.401 & 0.606 & 0.465 & 0.525 & 0.726 & 0.686 & 0.568 \\
        Macenko                      & Input                & 0.306 & 0.391 & 0.533 & 0.511 & 0.410 & 0.471 & 0.437 \\
        PCA projection               & Input                & 0.288 & 0.566 & 0.669 & 0.644 & 0.831 & 0.816 & 0.636 \\
        FIX                          & Input                & 0.605 & 0.335 & 0.689 & 0.781 & 0.881 & 0.844 & 0.689 \\
        \midrule
        MS                           & Feature              & 0.485 & 0.527 & 0.652 & 0.752 & 0.804 & 0.844 & 0.677 \\
        FS                           & Feature              & 0.523 & 0.533 & 0.668 & 0.753 & 0.792 & 0.846 & 0.686 \\
        CORAL                        & Feature              & 0.433 & 0.449 & 0.604 & 0.577 & 0.797 & 0.771 & 0.605 \\
        Ledoit--Wolf CORAL           & Feature              & 0.441 & 0.446 & 0.626 & 0.606 & 0.805 & 0.780 & 0.617 \\
        \midrule
        PCA projection + MS            & Input + feature      & 0.461 & 0.613 & 0.725 & 0.826 & 0.856 & 0.892 & 0.729 \\
        FIX + MS                       & Input + feature      & 0.567 & 0.501 & 0.755 & 0.826 & 0.873 & 0.881 & 0.734 \\
        FIX + FS                       & Input + feature      & 0.599 & 0.531 & 0.781 & 0.820 & 0.873 & 0.881 & \textbf{0.748}\\
        \bottomrule
    \end{tabular}
\end{table*}

\paragraph{Input-space correction results.}
Table~\ref{tab:correction_uni} shows the tile-level results for UNI, where FIX achieved the highest mean macro-F1 among the standalone
input-space methods (0.689), followed by PCA projection (0.636), Reinhard
normalization (0.568), and Macenko normalization (0.437). FIX produced the
highest input-space result in five of the six transfer directions. The
exception was HER2 transfer from BCI to Institution A, where FIX decreased
macro-F1 to 0.335, compared with 0.564 without correction. One possible explanation is that the target cohort contained only 15 patients, leaving few patients in each calibration subset.
Table~\ref{tab:correction_v2} shows the tile-level results for Virchow2, where FIX again achieved the highest mean macro-F1 among the
standalone input-space methods (0.609), compared with 0.534 for PCA projection,
0.519 for Reinhard normalization, and 0.459 for Macenko normalization. FIX
produced the highest input-space result in four of the six transfer directions,
but not in either HER2 direction. Thus, FIX was the strongest input-space
method on average across the two encoders, although its benefit remained
dependent on the marker, transfer direction, and encoder.

\paragraph{Feature-space correction results.}
For UNI,  FS achieved the highest mean macro-F1 among the standalone
feature-space methods (0.686), followed by MS (0.677), the Ledoit--Wolf
shrinkage variant of CORAL (0.617), and CORAL (0.605) (Table~\ref{tab:correction_uni}). The small difference
between FS and MS indicates that mean alignment accounted for much of the
feature-space benefit, while matching the variance of each feature dimension
provided a modest additional gain on average. For Virchow2, in Table~\ref{tab:correction_v2}, FS again achieved the highest mean macro-F1 (0.677), compared
with 0.665 for MS, 0.641 for the shrinkage variant of CORAL, and 0.637 for
CORAL. FS estimates only a mean and variance for each feature dimension,
whereas CORAL estimates and aligns a full covariance matrix. Estimating the
full covariance can be less stable for the high-dimensional FM embeddings,
particularly with limited target calibration data. The slightly better result
of the shrinkage variant than standard CORAL is consistent with this
interpretation.

\paragraph{Combining input- and feature-space methods.}
Each cross-space combination achieved a higher overall mean macro-F1 than both of its corresponding standalone methods. For UNI, in Table~\ref{tab:correction_uni}, 
PCA projection+MS achieved a
mean macro-F1 of 0.729, compared with 0.636 for PCA projection and 0.677 for
MS. Similarly, FIX+MS achieved 0.734, compared with 0.689 for FIX and 0.677
for MS. Combining the strongest standalone input-space method with the strongest
standalone feature-space method produced the numerically highest overall
performance. FIX+FS achieved 0.748, compared with 0.689 for FIX and 0.686 for
FS. Similar trends were seen for Virchow2 in Table~\ref{tab:correction_v2}, where FIX+FS achieved 0.738, compared with 0.609 for FIX and 0.677 for FS. These results indicate that input- and feature-space corrections provided
complementary benefits on average.

\begin{table*}[t]
    \centering
    \small
    \setlength{\tabcolsep}{3pt}
    \renewcommand{\arraystretch}{1.1}
\caption{Tile-level macro-F1 for label-free correction methods using
Virchow2. Values are means over five target calibration--evaluation splits.
Mean denotes the average across the six transfer directions. Bold indicates the numerically highest value.}
    \label{tab:correction_v2}
    \begin{tabular}{@{}llccccccc@{}}
        \toprule
        \multirow{2}{*}{Method}
        & \multirow{2}{*}{Correction stage}
        & \multicolumn{2}{c}{HER2}
        & \multicolumn{2}{c}{Nectin-4}
        & \multicolumn{2}{c}{TROP2} \\
        \cmidrule(lr){3-4}
        \cmidrule(lr){5-6}
        \cmidrule(lr){7-8}
        & &
        $\mathrm{A}\to\mathrm{BCI}$ &
        $\mathrm{BCI}\to\mathrm{A}$ &
        $\mathrm{A}\to\mathrm{B}$ &
        $\mathrm{B}\to\mathrm{A}$ &
        $\mathrm{A}\to\mathrm{B}$ &
        $\mathrm{B}\to\mathrm{A}$ &
        Mean \\
        \midrule
        No correction                & None                 & 0.578 & 0.170 & 0.526 & 0.550 & 0.893 & 0.461 & 0.530 \\
        \midrule
        Reinhard                     & Input                & 0.633 & 0.375 & 0.385 & 0.480 & 0.752 & 0.489 & 0.519 \\
        Macenko                      & Input                & 0.501 & 0.409 & 0.270 & 0.505 & 0.698 & 0.367 & 0.459 \\
        PCA projection               & Input                & 0.397 & 0.147 & 0.598 & 0.666 & 0.781 & 0.616 & 0.534 \\
        FIX                          & Input                & 0.625 & 0.153 & 0.605 & 0.750 & 0.847 & 0.676 & 0.609 \\
        \midrule
        MS                           & Feature              & 0.584 & 0.522 & 0.627 & 0.757 & 0.823 & 0.674 & 0.665 \\
        FS                           & Feature              & 0.585 & 0.574 & 0.630 & 0.765 & 0.819 & 0.687 & 0.677 \\
        CORAL                        & Feature              & 0.553 & 0.583 & 0.589 & 0.572 & 0.791 & 0.732 & 0.637 \\
        Ledoit--Wolf CORAL           & Feature              & 0.556 & 0.563 & 0.598 & 0.594 & 0.797 & 0.737 & 0.641 \\
        \midrule
        PCA projection + MS            & Input + feature      & 0.530 & 0.388 & 0.757 & 0.765 & 0.897 & 0.810 & 0.691 \\
        FIX + MS                       & Input + feature      & 0.681 & 0.481 & 0.749 & 0.802 & 0.888 & 0.819 & 0.737 \\
        FIX + FS                       & Input + feature      & 0.694 & 0.483 & 0.761 & 0.785 & 0.887 & 0.821 & \textbf{0.738} \\
        \bottomrule
    \end{tabular}
\end{table*}


\begin{table}[t]
\centering
\small
\caption{Patient-averaged tile-level performance. 
Each identified patient received equal total weight. For BCI, released
original-image identifiers served as grouping units because patient
identifiers were unavailable. Values are means across the 12 settings.}
\label{tab:patient_weighted}
\begin{tabular}{lccc}
\toprule
Method & Macro-F1 $\uparrow$ & QWK $\uparrow$ & Ordinal error $\downarrow$ \\
\midrule
No correction & 0.539 & 0.536 & 0.478 \\
FIX           & 0.652 & 0.673 & 0.332 \\
FS            & 0.662 & 0.675 & 0.329 \\
FIX + FS        & \textbf{0.719} & \textbf{0.744} & \textbf{0.269} \\
\bottomrule
\end{tabular}
\end{table}
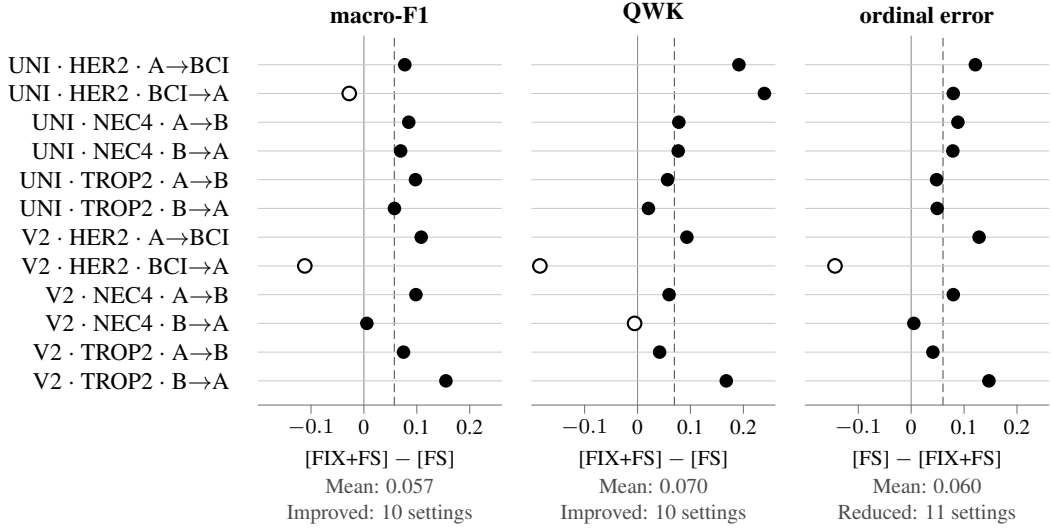
\begin{figure}[t]
\centering
\begin{tikzpicture}[font=\small, line cap=round]


\node[anchor=south, font=\small\bfseries] at (5.210,0.420) {macro-F1};
\draw[black!45, thin] (5.000,-4.500) -- (5.000,0.380);
\draw[black!70, densely dashed, thin] (5.402,-4.500) -- (5.402,0.380);
\draw[black!55, thin] (3.600,-4.500) -- (6.820,-4.500);
\draw[black!55, thin] (4.300,-4.500) -- (4.300,-4.610);
\node[anchor=north, font=\footnotesize, inner sep=2pt] at (4.300,-4.610) {$-0.1$};
\draw[black!55, thin] (5.000,-4.500) -- (5.000,-4.610);
\node[anchor=north, font=\footnotesize, inner sep=2pt] at (5.000,-4.610) {0};
\draw[black!55, thin] (5.700,-4.500) -- (5.700,-4.610);
\node[anchor=north, font=\footnotesize, inner sep=2pt] at (5.700,-4.610) {0.1};
\draw[black!55, thin] (6.400,-4.500) -- (6.400,-4.610);
\node[anchor=north, font=\footnotesize, inner sep=2pt] at (6.400,-4.610) {0.2};
\node[anchor=north, font=\footnotesize, inner sep=2pt] at (5.210,-5.020) {[FIX+FS] $-$ [FS]};
\node[anchor=north, font=\footnotesize, black!70, inner sep=2pt] at (5.210,-5.400) {Mean: 0.057};
\node[anchor=north, font=\footnotesize, black!70, inner sep=2pt] at (5.210,-5.740) {Improved: 10 settings};

\node[anchor=south, font=\small\bfseries] at (8.830,0.420) {QWK};
\draw[black!45, thin] (8.620,-4.500) -- (8.620,0.380);
\draw[black!70, densely dashed, thin] (9.107,-4.500) -- (9.107,0.380);
\draw[black!55, thin] (7.220,-4.500) -- (10.440,-4.500);
\draw[black!55, thin] (7.920,-4.500) -- (7.920,-4.610);
\node[anchor=north, font=\footnotesize, inner sep=2pt] at (7.920,-4.610) {$-0.1$};
\draw[black!55, thin] (8.620,-4.500) -- (8.620,-4.610);
\node[anchor=north, font=\footnotesize, inner sep=2pt] at (8.620,-4.610) {0};
\draw[black!55, thin] (9.320,-4.500) -- (9.320,-4.610);
\node[anchor=north, font=\footnotesize, inner sep=2pt] at (9.320,-4.610) {0.1};
\draw[black!55, thin] (10.020,-4.500) -- (10.020,-4.610);
\node[anchor=north, font=\footnotesize, inner sep=2pt] at (10.020,-4.610) {0.2};
\node[anchor=north, font=\footnotesize, inner sep=2pt] at (8.830,-5.020) {[FIX+FS] $-$ [FS]};
\node[anchor=north, font=\footnotesize, black!70, inner sep=2pt] at (8.830,-5.400) {Mean: 0.070};
\node[anchor=north, font=\footnotesize, black!70, inner sep=2pt] at (8.830,-5.740) {Improved: 10 settings};

\node[anchor=south, font=\small\bfseries] at (12.450,0.420) {ordinal error};
\draw[black!45, thin] (12.240,-4.500) -- (12.240,0.380);
\draw[black!70, densely dashed, thin] (12.661,-4.500) -- (12.661,0.380);
\draw[black!55, thin] (10.840,-4.500) -- (14.060,-4.500);
\draw[black!55, thin] (11.540,-4.500) -- (11.540,-4.610);
\node[anchor=north, font=\footnotesize, inner sep=2pt] at (11.540,-4.610) {$-0.1$};
\draw[black!55, thin] (12.240,-4.500) -- (12.240,-4.610);
\node[anchor=north, font=\footnotesize, inner sep=2pt] at (12.240,-4.610) {0};
\draw[black!55, thin] (12.940,-4.500) -- (12.940,-4.610);
\node[anchor=north, font=\footnotesize, inner sep=2pt] at (12.940,-4.610) {0.1};
\draw[black!55, thin] (13.640,-4.500) -- (13.640,-4.610);
\node[anchor=north, font=\footnotesize, inner sep=2pt] at (13.640,-4.610) {0.2};
\node[anchor=north, font=\footnotesize, inner sep=2pt] at (12.450,-5.020) {[FS] $-$ [FIX+FS]};
\node[anchor=north, font=\footnotesize, black!70, inner sep=2pt] at (12.450,-5.400) {Mean: 0.060};
\node[anchor=north, font=\footnotesize, black!70, inner sep=2pt] at (12.450,-5.740) {Reduced: 11 settings};

\node[anchor=east, font=\small, inner sep=2pt] at (3.300,-0.000) {UNI $\cdot$ HER2 $\cdot$ A$\to$BCI};
\draw[black!20, thin] (3.600,-0.000) -- (6.820,-0.000);
\node[circle, fill=black, inner sep=1.8pt] at (5.541,-0.000) {};
\draw[black!20, thin] (7.220,-0.000) -- (10.440,-0.000);
\node[circle, fill=black, inner sep=1.8pt] at (9.961,-0.000) {};
\draw[black!20, thin] (10.840,-0.000) -- (14.060,-0.000);
\node[circle, fill=black, inner sep=1.8pt] at (13.089,-0.000) {};

\node[anchor=east, font=\small, inner sep=2pt] at (3.300,-0.380) {UNI $\cdot$ HER2 $\cdot$ BCI$\to$A};
\draw[black!20, thin] (3.600,-0.380) -- (6.820,-0.380);
\node[circle, fill=white, draw=black, thick, inner sep=1.8pt] at (4.806,-0.380) {};
\draw[black!20, thin] (7.220,-0.380) -- (10.440,-0.380);
\node[circle, fill=black, inner sep=1.8pt] at (10.298,-0.380) {};
\draw[black!20, thin] (10.840,-0.380) -- (14.060,-0.380);
\node[circle, fill=black, inner sep=1.8pt] at (12.797,-0.380) {};

\node[anchor=east, font=\small, inner sep=2pt] at (3.300,-0.760) {UNI $\cdot$ NEC4 $\cdot$ A$\to$B};
\draw[black!20, thin] (3.600,-0.760) -- (6.820,-0.760);
\node[circle, fill=black, inner sep=1.8pt] at (5.593,-0.760) {};
\draw[black!20, thin] (7.220,-0.760) -- (10.440,-0.760);
\node[circle, fill=black, inner sep=1.8pt] at (9.166,-0.760) {};
\draw[black!20, thin] (10.840,-0.760) -- (14.060,-0.760);
\node[circle, fill=black, inner sep=1.8pt] at (12.858,-0.760) {};

\node[anchor=east, font=\small, inner sep=2pt] at (3.300,-1.140) {UNI $\cdot$ NEC4 $\cdot$ B$\to$A};
\draw[black!20, thin] (3.600,-1.140) -- (6.820,-1.140);
\node[circle, fill=black, inner sep=1.8pt] at (5.486,-1.140) {};
\draw[black!20, thin] (7.220,-1.140) -- (10.440,-1.140);
\node[circle, fill=black, inner sep=1.8pt] at (9.158,-1.140) {};
\draw[black!20, thin] (10.840,-1.140) -- (14.060,-1.140);
\node[circle, fill=black, inner sep=1.8pt] at (12.791,-1.140) {};

\node[anchor=east, font=\small, inner sep=2pt] at (3.300,-1.520) {UNI $\cdot$ TROP2 $\cdot$ A$\to$B};
\draw[black!20, thin] (3.600,-1.520) -- (6.820,-1.520);
\node[circle, fill=black, inner sep=1.8pt] at (5.682,-1.520) {};
\draw[black!20, thin] (7.220,-1.520) -- (10.440,-1.520);
\node[circle, fill=black, inner sep=1.8pt] at (9.017,-1.520) {};
\draw[black!20, thin] (10.840,-1.520) -- (14.060,-1.520);
\node[circle, fill=black, inner sep=1.8pt] at (12.576,-1.520) {};

\node[anchor=east, font=\small, inner sep=2pt] at (3.300,-1.900) {UNI $\cdot$ TROP2 $\cdot$ B$\to$A};
\draw[black!20, thin] (3.600,-1.900) -- (6.820,-1.900);
\node[circle, fill=black, inner sep=1.8pt] at (5.404,-1.900) {};
\draw[black!20, thin] (7.220,-1.900) -- (10.440,-1.900);
\node[circle, fill=black, inner sep=1.8pt] at (8.764,-1.900) {};
\draw[black!20, thin] (10.840,-1.900) -- (14.060,-1.900);
\node[circle, fill=black, inner sep=1.8pt] at (12.584,-1.900) {};

\node[anchor=east, font=\small, inner sep=2pt] at (3.300,-2.280) {V2 $\cdot$ HER2 $\cdot$ A$\to$BCI};
\draw[black!20, thin] (3.600,-2.280) -- (6.820,-2.280);
\node[circle, fill=black, inner sep=1.8pt] at (5.758,-2.280) {};
\draw[black!20, thin] (7.220,-2.280) -- (10.440,-2.280);
\node[circle, fill=black, inner sep=1.8pt] at (9.273,-2.280) {};
\draw[black!20, thin] (10.840,-2.280) -- (14.060,-2.280);
\node[circle, fill=black, inner sep=1.8pt] at (13.138,-2.280) {};

\node[anchor=east, font=\small, inner sep=2pt] at (3.300,-2.660) {V2 $\cdot$ HER2 $\cdot$ BCI$\to$A};
\draw[black!20, thin] (3.600,-2.660) -- (6.820,-2.660);
\node[circle, fill=white, draw=black, thick, inner sep=1.8pt] at (4.218,-2.660) {};
\draw[black!20, thin] (7.220,-2.660) -- (10.440,-2.660);
\node[circle, fill=white, draw=black, thick, inner sep=1.8pt] at (7.328,-2.660) {};
\draw[black!20, thin] (10.840,-2.660) -- (14.060,-2.660);
\node[circle, fill=white, draw=black, thick, inner sep=1.8pt] at (11.232,-2.660) {};

\node[anchor=east, font=\small, inner sep=2pt] at (3.300,-3.040) {V2 $\cdot$ NEC4 $\cdot$ A$\to$B};
\draw[black!20, thin] (3.600,-3.040) -- (6.820,-3.040);
\node[circle, fill=black, inner sep=1.8pt] at (5.688,-3.040) {};
\draw[black!20, thin] (7.220,-3.040) -- (10.440,-3.040);
\node[circle, fill=black, inner sep=1.8pt] at (9.037,-3.040) {};
\draw[black!20, thin] (10.840,-3.040) -- (14.060,-3.040);
\node[circle, fill=black, inner sep=1.8pt] at (12.797,-3.040) {};

\node[anchor=east, font=\small, inner sep=2pt] at (3.300,-3.420) {V2 $\cdot$ NEC4 $\cdot$ B$\to$A};
\draw[black!20, thin] (3.600,-3.420) -- (6.820,-3.420);
\node[circle, fill=black, inner sep=1.8pt] at (5.039,-3.420) {};
\draw[black!20, thin] (7.220,-3.420) -- (10.440,-3.420);
\node[circle, fill=white, draw=black, thick, inner sep=1.8pt] at (8.581,-3.420) {};
\draw[black!20, thin] (10.840,-3.420) -- (14.060,-3.420);
\node[circle, fill=black, inner sep=1.8pt] at (12.275,-3.420) {};

\node[anchor=east, font=\small, inner sep=2pt] at (3.300,-3.800) {V2 $\cdot$ TROP2 $\cdot$ A$\to$B};
\draw[black!20, thin] (3.600,-3.800) -- (6.820,-3.800);
\node[circle, fill=black, inner sep=1.8pt] at (5.524,-3.800) {};
\draw[black!20, thin] (7.220,-3.800) -- (10.440,-3.800);
\node[circle, fill=black, inner sep=1.8pt] at (8.912,-3.800) {};
\draw[black!20, thin] (10.840,-3.800) -- (14.060,-3.800);
\node[circle, fill=black, inner sep=1.8pt] at (12.527,-3.800) {};

\node[anchor=east, font=\small, inner sep=2pt] at (3.300,-4.180) {V2 $\cdot$ TROP2 $\cdot$ B$\to$A};
\draw[black!20, thin] (3.600,-4.180) -- (6.820,-4.180);
\node[circle, fill=black, inner sep=1.8pt] at (6.085,-4.180) {};
\draw[black!20, thin] (7.220,-4.180) -- (10.440,-4.180);
\node[circle, fill=black, inner sep=1.8pt] at (9.794,-4.180) {};
\draw[black!20, thin] (10.840,-4.180) -- (14.060,-4.180);
\node[circle, fill=black, inner sep=1.8pt] at (13.268,-4.180) {};

\end{tikzpicture}

\caption{Per-setting effect of adding FIX to FS, under patient-averaged tile-level evaluation. Each point shows the difference between FIX+FS and FS
in the mean over the five calibration--evaluation splits, using the same
weighting as Table~\ref{tab:patient_weighted}. Ordinal error is
lower-is-better, so its panel plots FS $-$ FIX+FS; positive values therefore
indicate better performance with FIX+FS in all three panels. The three panels
share the same horizontal scale. The solid vertical line marks zero, and the
dashed line marks the mean across the 12 settings. Open circles indicate
settings in which FIX+FS performs worse than FS. V2 denotes Virchow2 and NEC4
denotes Nectin-4.}
\label{fig:paired}

\end{figure}

\paragraph{Additional analyses.}
Table~\ref{tab:patient_weighted} reports performance analysis in which
each identified patient received equal total weight.
FIX+FS achieved the highest mean
patient-averaged tile-level performance, with a macro-F1 of 0.719, QWK of
0.744, and ordinal error of 0.269. Relative to no correction, FIX+FS improved
macro-F1 and QWK by 0.181 and 0.208, respectively, and reduced ordinal error
by 0.209. 
The corresponding improvements occurred in 10/12, 11/12, and 11/12
settings (descriptive two-sided sign tests: $p=0.039$, $p=0.006$, and
$p=0.006$, respectively). Under these descriptive tests, FIX+FS performed better than no correction
across settings for all three metrics.
Figure~\ref{fig:paired} compares FIX+FS with FS to isolate the observed contribution of adding FIX to
the strongest standalone feature-space method. FIX+FS improved macro-F1 in 10
of the 12 settings, with a mean difference of 0.057 (paired Wilcoxon
signed-rank $p=0.042$ over the 12 setting-level differences). The
corresponding QWK and ordinal-error differences were 0.070 and 0.060, with
improvements in 10 and 11 settings. Because the 12 settings reuse cohorts and
images, this $p$-value is descriptive rather than confirmatory.

\section{Discussion}

The results support the complementary use of input- and feature-space
corrections for cross-institutional IHC grading with frozen pathology FMs.
FIX and FS were the strongest standalone methods at their respective stages,
and their combination achieved the highest overall performance. Importantly, the same ranking was preserved in the patient-averaged tile-level analysis,
indicating that the observed benefit was not driven solely by differences in
the number of tiles contributed by each patient.
The two corrections address distribution shift at different points in the
pipeline. FIX modifies target images before frozen-FM encoding, whereas FS
aligns the resulting feature dimensions used for probe training. Their
combination can therefore correct discrepancies that may not be adequately
addressed in either space alone. FS also performed better on average than
full covariance alignment, suggesting that simple dimension-wise alignment
may be effective when calibration data are limited relative to the feature
dimension. Together, these findings demonstrate a practical, label-free
strategy for improving cross-institutional transfer without updating the
foundation model or using target evaluation labels.


\paragraph{Limitations.} This study included three IHC markers, two frozen encoders, and a limited
number of institution pairs; newer pathology FMs such as UNI2
\citep{mahmoodlab2025uni2} and GenBio-PathFM
\citep{kapse2026genbiopathfm} were not evaluated. The 12
settings reuse cohorts and images and therefore do not constitute independent
external validations. BCI released
original-image identifiers were used as proxies for patients because patient identifiers were unavailable. Finally, FIX's gating
constants were not validated on a separate IHC marker or institution. Future
work should evaluate newer encoders and validate the method
across additional institutions, biomarkers, and naturally sampled
populations.

\section{Conclusion}
In this work, we explored label-free corrections applied before and after frozen-FM encoding for cross-institutional IHC grading. For a total of 12 different settings, we show that FIX, our proposed method, achieved the
highest mean macro-F1 among the evaluated standalone input-space methods, while
FS achieved the highest mean among the feature-space methods. Their combination,
FIX+FS, produced the highest overall mean macro-F1 and QWK and the lowest
ordinal error. These findings
show that combining input- and feature-space corrections results in complementary benefits.

\section*{Ethics and Data Availability}
The two in-house cohorts were obtained from Institutions A and B. The study was
approved by the institutional review board of each participating institution,
with a waiver of informed consent for retrospective analysis. The third cohort is the
public BCI dataset~\citep{liu2022bci}.
The public cohort is used under its original release terms and is
cited accordingly~\citep{liu2022bci}. Code for the correction stages,
evaluation protocol, and aggregation scripts will be released at
\url{https://jongwoonalee.github.io}.

\begin{ack}
We thank the pathology departments and research teams at the participating
institutions for their cooperation with this study. This work received no
external funding.
\end{ack}

\bibliographystyle{plainnat}
\bibliography{refs}


\appendix

\setcounter{section}{0}
\setcounter{table}{0}
\setcounter{figure}{0}
\setcounter{equation}{0}
\renewcommand{\thesection}{S\arabic{section}}
\renewcommand{\thetable}{S\arabic{table}}
\renewcommand{\thefigure}{S\arabic{figure}}
\renewcommand{\theequation}{S\arabic{equation}}

\clearpage

\section{Implementation details}
\label{sup:implemenation_details}

\subsection{FIX component-wise gating}
\label{sup:fix_gating}

FIX does not correct every PCA component equally. It first measures how
different the source and target calibration patches are along each component.
If the differences are concentrated in only a few components, FIX focuses the
correction on those components. If the differences are spread across many
components, it applies smaller corrections more broadly. Components showing
little source--target difference receive little or no correction. The
following equations define how FIX selects among these correction patterns
using source and unlabeled target calibration data only.

The per-component strengths derive from the drift score $r$ and a global
strength fixed at $\alpha=1.0$ for every marker, encoder, transfer direction,
and split. Let $\sigma(\cdot)$ be the logistic function, $\tilde r$ the median
drift score, $\mathrm{IQR}(r)$ its interquartile range, $\lambda_i$ the
eigenvalue of component $i$, $\gamma=3$, and $\varepsilon_g=10^{-6}$. Four
branches exist in the implementation:
\begin{align}
\text{thresholded:}\quad &\alpha_i=\alpha\,\sigma\!\big(\gamma(r_i-1)\big)\,
  \mathbf{1}[r_i>1],\\
\text{continuous:}\quad &\alpha_i=\alpha\,
  \sigma\!\Big(\gamma\tfrac{r_i-\tilde r}{\mathrm{IQR}(r)+\varepsilon_g}\Big),\\
\text{weak continuous:}\quad &\alpha_i=\tfrac{1}{2}\alpha\,
  \sigma\!\Big(\gamma\tfrac{r_i-\tilde r}{\mathrm{IQR}(r)+\varepsilon_g}\Big),\\
\text{cumulative drift:}\quad &\alpha_i=\alpha\,\mathbf{1}[i\in\mathcal{M}],
\end{align}
all clipped to $[0,1]$. If $\mathrm{IQR}(r)<10^{-6}$, the standard deviation of
$r$ replaces the interquartile range; the same $\varepsilon_g=10^{-6}$ is then
added to the resulting scale. The set $\mathcal{M}$ is the smallest set of
components, ordered by decreasing $r$, whose cumulative eigenvalue-weighted
drift share reaches $\tau=0.9$.

The branch is selected automatically from the drift profile and the source PCA
eigenvalues, without using target labels or target evaluation performance. With
$w_i\propto r_i\lambda_i$, $n_{\mathrm{high}}=|\{i:r_i>1\}|$, and $m_{90}$ the
number of components needed to reach $\tau$ of $\sum_i w_i$:

\begin{itemize}
\item $n_{\mathrm{high}}=0$: weak continuous if $\sum_i r_i>0.05$, otherwise
  $\alpha\equiv0$;
\item $n_{\mathrm{high}}\le0.15K$ and $\sum_{r_i>1}w_i/\sum_i w_i\ge0.60$:
  thresholded;
\item $m_{90}\le0.60K$: cumulative drift;
\item otherwise: continuous.
\end{itemize}

In the primary experiments, only the thresholded and weak-continuous branches
were selected. The thresholded branch applied to HER2 alone, activating 7 to 8
of the 64 components at a mean strength near 0.87; Nectin-4 and TROP2 used weak
continuous rescaling across all 64 components at a mean strength of 0.26. The
gating constants were not validated on a separate marker or institution. A
held-out marker or institution would provide stronger validation and was not
evaluated here.

\subsection{Data details}
\label{sup:data_details}

\paragraph{Incomplete patient identifiers}
BCI does not provide patient identifiers; however, it contains 984 released original-image identifiers, which we used as grouping units (pseudo-patient identifier) for metrics and calibration set splits. Nectin-4 patient identifiers were available for 1{,}023 of the 1{,}500 Institution A
Nectin-4 tiles (68.2\%), representing 82 identified patients; the remaining
477 tiles (31.8\%) could not be linked to patients. When Institution A served
as the target, the primary analysis used only the 1{,}023 resolved tiles so
that patient-level separation between calibration and evaluation could be
enforced.

\paragraph{Target Calibration set splits}
Target tiles are split into a calibration set and an evaluation set. The calibration set is used to estimate each correction method's parameters and statistics. For example, mean-shift (MS) uses the calibration set to calculate the target feature mean. During the split, tiles from the same patient are kept on the same side of each calibration--evaluation split.  Patients are assigned to the calibration and evaluation sides by a greedy rule
that keeps the three-grade tile distribution of the two sides as close as
possible. Every grade is present on both sides in all sixty splits. The split is approximately a 50/50 split.

\paragraph{Automated tissue screening.}
A tile was excluded as non-tissue when at least 50\% of its pixels were
background, a pixel counting as background when its mean RGB intensity exceeded
220. Depending on marker and site, this prespecified rule excluded 46--77\% of
candidate tiles. No further tile-level exclusion was applied.

\begin{table*}[t]
    \centering
    \caption{
    Training and evaluation procedures for the evaluated conditions. All conditions
    use the same frozen pathology FM and the same source grades. Held-out target
    evaluation data are used neither for correction estimation nor for probe
    training.
    }
    \label{tab:conditions}
    
    \small
    \setlength{\tabcolsep}{4pt}
    \renewcommand{\arraystretch}{1.3}
    
    \begin{tabular}{
    p{0.14\textwidth}
    p{0.26\textwidth}
    p{0.28\textwidth}
    p{0.22\textwidth}
    }
    \toprule
    \textbf{Condition}
    &
    \textbf{Target-calibration use}
    &
    \textbf{Source / probe-training representation}
    &
    \textbf{Held-out target representation}
    \\
    \midrule
    
    No correction
    &
    None
    &
    Unmodified source images encoded by the frozen FM
    &
    Unmodified target images
    \\
    \addlinespace
    
    PCA projection
    &
    None; the basis is fitted on source patches only
    &
    PCA-reconstructed source images encoded by the frozen FM
    &
    PCA-reconstructed target images
    \\
    \addlinespace
    
    FIX
    &
    Estimate coefficient statistics from source and unlabeled target calibration
    RGB patches
    &
    PCA-reconstructed source images, without coefficient editing
    &
    FIX-corrected target images
    \\
    \addlinespace
    
    MS
    &
    Mean of the unmodified target calibration features
    &
    Unmodified source features recentered to that mean
    &
    Unmodified target images
    \\
    \addlinespace
    
    FS
    &
    Per-dimension mean and standard deviation of the unmodified target calibration
    features
    &
    Unmodified source features standardized to those statistics
    &
    Unmodified target images
    \\
    \addlinespace
    
    PCA projection+MS
    &
    Mean of the PCA-reconstructed target calibration features
    &
    PCA-reconstructed source features recentered to that mean
    &
    PCA-reconstructed target images, unshifted
    \\
    \addlinespace
    
    FIX+MS
    &
    Estimate FIX, then take the mean of the FIX-corrected target calibration
    features
    &
    PCA-reconstructed source features mean-aligned to that mean
    &
    FIX-corrected target images, unshifted
    \\
    \addlinespace
    
    FIX+FS
    &
    Estimate FIX, then take the per-dimension mean and standard deviation of the
    FIX-corrected target calibration features
    &
    PCA-reconstructed source features feature-wise standardized to those statistics
    &
    FIX-corrected target images, untransformed after encoding
    \\
    
    \bottomrule
    \end{tabular}
\end{table*}

\subsection{Source, calibration, and held-out target dataset usage}
\label{sup:source_calibration_held_out}

Table~\ref{tab:conditions} summarizes how each dataset is used on each method.

\end{document}